\documentclass[runningheads]{llncs}
\usepackage[T1]{fontenc}
\usepackage{graphicx}
\usepackage{booktabs}
\usepackage[misc]{ifsym}
\newcommand{\corr}{(\Letter)}
\usepackage{mwe}
\usepackage{amsmath}
\usepackage{amssymb}
\usepackage{amsfonts}
\usepackage{multirow}
\usepackage{tikz}
\usepackage{pgfplots}
\usepackage{pgfplotstable}
\pgfplotsset{compat=1.18}
\usepgfplotslibrary{groupplots}
\usepackage[table]{xcolor}

\begin{document}

\title{Does Role Specialization Matter for Explanation Faithfulness in Mixture-of-Experts?}

\titlerunning{Does Role Specialization Matter for Explanation Faithfulness in MoE?}

\author{Yeji Kim\inst{1}\corr \and
Housam Babiker\inst{2} \and
Mi-Young Kim\inst{3} \and
Randy Goebel\inst{1}}
\authorrunning{Y. Kim et al.}

\institute{University of Alberta, Edmonton, AB, Canada\\ 
\email{\{yeji7,rgoebel\}@ualberta.ca}
\and
University of Waikato, Hamilton, New Zealand\\ \email{housam.babiker@waikato.ac.nz}
\and
University of Alberta, Camrose, AB, Canada\\
\email{miyoung2@ualberta.ca}}

\toctitle{Does Role Specialization Matter for Explanation Faithfulness in Mixture-of-Experts?}
\tocauthor{Yeji Kim, Housam Babiker, Mi-Young Kim, Randy Goebel}
\maketitle              

\begin{abstract}
Mixture-of-Experts (MoE) architectures have recently been extended with role-based mechanisms for interpretability. This is typically done by assigning semantic roles to individual expert components, for example roles like synergy, redundancy, and uniqueness in multimodal settings. However, whether such structural role decomposition preserves explanation faithfulness of the overall architecture remains largely underexplored. We hypothesize that inter-expert representation overlap weakens effective role separation and degrades attribution-based faithfulness, even when semantic roles are explicitly defined. 

To address this limitation, we introduce representation-level decorrelation regularization to explicitly reduce inter-expert similarity in latent space. Using representation decorrelation objectives, we encourage clearer specialization among experts by minimizing representation overlap. Our experiments show that across multiple multimodal benchmarks, this separation consistently improves explanation faithfulness, as measured by comprehensiveness, sufficiency, and their Area Over the Perturbation Curve (AOPC) summaries, while preserving task performance. 

We further show that these improvements are not limited to role-based architectures such as Interpretable Multimodal Interaction-aware MoE (I$^2$MoE). Similar trends are observed in a standard sparse MoE baseline, suggesting that representation-level separation may provide a more general mechanism for enhancing explanation faithfulness in MoE systems. Overall, our findings suggest that structural role decomposition alone may be insufficient to guarantee faithful explanations and that representation-level separation helps improve explanation faithfulness. To support reproducibility, the source code and supplementary material are publicly available at https://github.com/dut0817/FL-I2MoE\_Decor.

\end{abstract}

\section{Introduction}
Mixture-of-Experts (MoE) architectures enable conditional computation by activating only a subset of expert modules per input, allowing models to scale capacity without proportional increases in computation~\cite{shazeer2017outrageously,lepikhin2020gshard,fedus2022switch}. A key property of MoE models is expert specialization. Through routing mechanisms, different experts tend to capture distinct patterns or sub-functions of a task~\cite{jacobs1991adaptive,lo2025closer}. This property is particularly appealing in multimodal learning, where heterogeneous signals across modalities often require specialized processing. Motivated by this observation, recent multimodal MoE architectures explicitly assign semantic roles to experts, including uniqueness, redundancy, and synergy, in order to decompose cross-modal interactions into interpretable components~\cite{yu-etal-2024-mmoe,pmlr-v267-xin25c}.

However, whether such structural role decomposition preserves the explanation faithfulness remains unclear. In practice, role-based experts may still learn highly overlapping representations, which results in correlated latent spaces and ambiguous specialization~\cite{do2025simsmoe,feng2025comoe}. This phenomenon partly arises from the design of MoE architectures. Experts typically receive representations produced by shared encoders and are optimized with the same task objective. Although routing determines which experts are activated for a given input, it does not explicitly enforce separation between expert representations~\cite{cai2025survey,chen2022towards}. As a result, experts that are assigned different semantic roles may still occupy overlapping regions of the representation space. When this occurs, role assignments may become nominal, and feature attributions may fail to reflect the actual evidence used by the MoE model's prediction. These observations raise an important question: does role specialization in MoE architectures truly improve explanation faithfulness?

In our work, we study this question from the perspective of inter-expert representation overlap. Rather than proposing a new MoE architecture, we treat representation overlap as a controllable factor and examine how it influences explanation faithfulness. Using a feature-level extension of I$^2$MoE (FL-I$^2$MoE)~\cite{pmlr-v267-xin25c} as the backbone, we manipulate overlap through representation-level decorrelation regularization while keeping the architecture, routing mechanism, and attribution procedure fixed. We consider three separation objectives that operate on different geometric properties of the representation space. These include cosine similarity minimization, Centered Kernel Alignment (CKA) decorrelation~\cite{kornblith2019similarity}, and a redundancy reduction objective inspired by Barlow Twins~\cite{zbontar2021barlow}.

Experiments across three multimodal benchmarks show that performance-preserving reductions in inter-expert representation overlap consistently improve perturbation-based faithfulness, including comprehensiveness, sufficiency, and their AOPC summaries~\cite{deyoung2020eraser}. In addition, representation geometry analyses confirm that the proposed regularization reduces similarity between expert representations, and synthetic experiments demonstrate improved alignment between routing behavior and intended expert roles. Similar trends also appear in a standard sparse MoE baseline without explicit semantic roles, which suggests that expert separation may improve explanation faithfulness beyond role-based architectures. 

Our contributions are summarized as follows.
\begin{itemize}
    \item We identify a gap between structural role decomposition and explanation faithfulness in role-based MoE architectures, and show that inter-expert representation overlap can weaken effective role specialization.
    \item We introduce representation-level decorrelation as a controlled mechanism for reducing inter-expert overlap while retaining the original architecture, routing mechanism, and attribution procedure.
    \item We demonstrate across multimodal benchmarks that performance-preserving reduced-overlap regimes improve explanation faithfulness, and that the same principle extends beyond explicitly role-based MoE architectures.
\end{itemize}

\section{Related Work}
\subsection{Mixture-of-Experts (MoE) Architectures}
MoE architectures scale neural networks through conditional computation by activating only a subset of experts for each input~\cite{shazeer2017outrageously,lepikhin2020gshard,fedus2022switch}. This design increases model predictive performance while keeping computational cost manageable. MoE models have also been increasingly explored in multimodal learning~\cite{lin2026moe,li2025uni}. 

\subsection{Role-based and Interaction-aware MoE}
Recent work extends MoE architectures beyond efficiency and explores their potential for interpretability in multimodal settings. Several approaches explicitly model different types of modality interactions. For example, MMoE~\cite{yu-etal-2024-mmoe} introduces interaction experts to capture redundancy, uniqueness, and synergy across modalities. Similarly, the Interpretable Multimodal Interaction-aware Mixture-of-Experts (I$^2$MoE)~\cite{pmlr-v267-xin25c} framework assigns semantic roles to experts and employs training objectives designed to align expert behavior with these interaction types. These approaches improve structural interpretability, but they do not explicitly enforce separation between expert representations. 

Although these architectures improve structural interpretability, their designs implicitly suggest  that assigning semantic roles is sufficient to induce meaningful specialization. In practice, however, experts associated with different roles may still learn highly overlapping representations, which can weaken effective specialization and limit the reliability of explanations.

\subsection{Representation Similarity and Expert Decorrelation}
Another line of work studies expert redundancy and representation collapse in MoE models. Prior studies have shown that experts can exhibit representation collapse or high similarity, reducing the effective capacity of the model~\cite{chi2022representation}. To address this issue, several methods encourage expert diversity during training. SimSMoE~\cite{do2025simsmoe} measures expert similarity using Centered Kernel Alignment (CKA) and applies regularization to reduce representation collapse. CoMoE~\cite{feng2025comoe} promotes expert specialization through contrastive objectives that increase representation separation. More generally, representation decorrelation has been widely studied in representation learning through cosine-based angular regularization, CKA-based similarity analysis, and Barlow Twins redundancy reduction~\cite{rodriguez2016regularizing,kornblith2019similarity,zbontar2021barlow}.

Despite these advances, prior work primarily studies representation separation from the perspective of optimization or representation learning. The relationship between expert representation overlap and explanation faithfulness in MoE architectures remains largely unexplored. In our work, we investigate whether reducing inter-expert representation similarity leads to clearer expert specialization and, consequently, more faithful feature-level explanations.

\section{Method}
\subsection{Feature-level Interaction-aware MoE Backbone}
\begin{figure}[t]
  \centering
  \includegraphics[width=\linewidth]{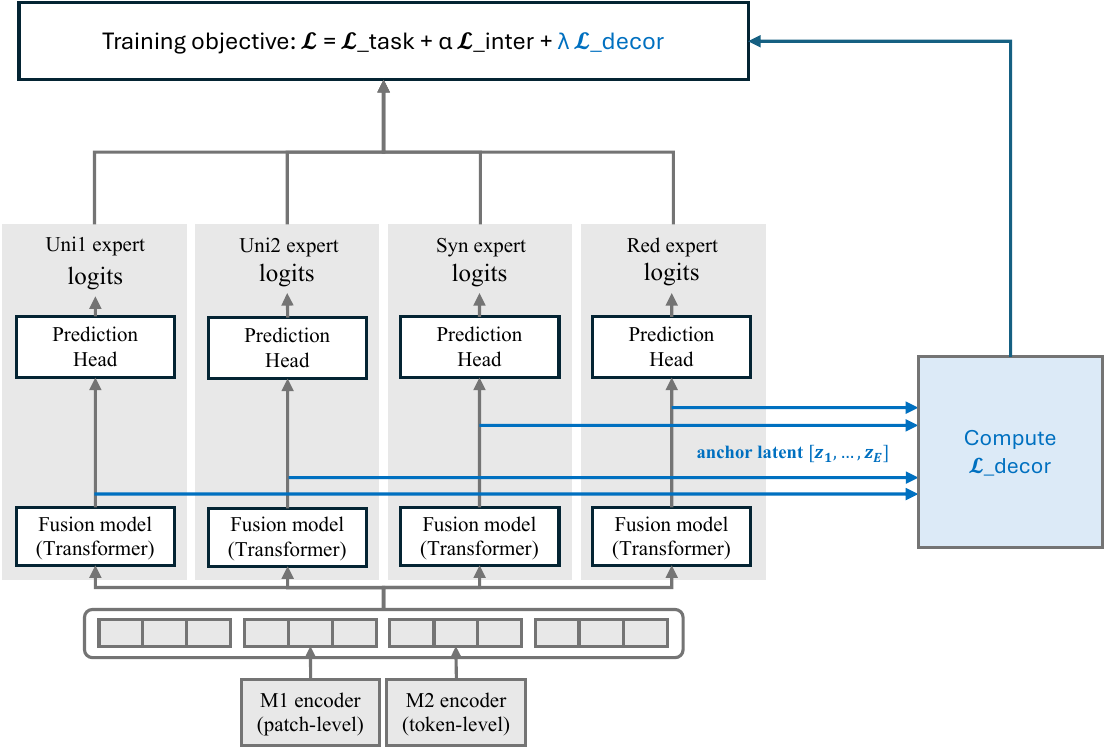}
  \caption{Overview of the proposed framework, illustrated for a bimodal setting. Shared modality encoders produce feature-level representations, which are passed to role-based interaction experts. Each expert contains its own fusion model and produces expert-specific logits and a latent representation. During training, the anchor latent representations are used to compute the representation-level decorrelation loss $\mathcal{L}_{\text{decor}}$, which is added to the original training objective.}
  \label{fig:overview}
\end{figure}

We use a feature-level variant of I$^2$MoE~\cite{pmlr-v267-xin25c}, denoted FL-I$^2$MoE~\cite{kim2026fli2moe}, as our backbone. Since I$^2$MoE explicitly imposes semantic roles on experts, it provides a controlled setting for isolating the effect of inter-expert representation overlap on explanation faithfulness without changing the underlying role-based MoE design. Unlike the original I$^2$MoE, which operates on pooled modality representations, FL-I$^2$MoE preserves feature-level representations so that attribution and perturbation can be applied directly to individual features. This provides us with the flexibility required for experimental design. Figure~\ref{fig:overview} provides an overview of the framework.

Given a multimodal input $x = \{x_1, \dots, x_M\}$ with $M$ modalities, each modality is encoded into a sequence of $d$-dimensional feature vectors
\begin{equation}
\phi_m(x_m) = \{h_{m,1}, \dots, h_{m,T_m}\}, 
\quad h_{m,t} \in \mathbb{R}^{d},
\end{equation} where $T_m$ is the number of features produced by the encoder for modality $m$.

The encoded modality features are shared across experts, \emph{but each interaction expert has its own fusion model}. The model contains $E=M+2$ experts: $M$ modality-specific uniqueness experts, one synergy expert modeling cross-modal interactions, and one redundancy expert capturing shared signals. Each expert produces class logits $\mathbf{y}_e(x)$ and a latent representation $\mathbf{z}_e(x) \in \mathbb{R}^{D}$. Expert logits are aggregated using routing weights $\mathbf{w}(x)$:
\begin{equation}
\hat{\mathbf{y}}(x) = \sum_{e=1}^{E} w_e(x)\,\mathbf{y}_e(x).
\end{equation}
These expert representations form the basis for analyzing inter-expert representation overlap.

\subsection{Role-based Interaction Learning}
Following I$^2$MoE~\cite{pmlr-v267-xin25c}, we retain the original role-specific interaction losses derived from Partial Information Decomposition (PID)~\cite{10.5555/3648699.3648830}. PID decomposes multimodal information into unique, redundant, and synergistic components, and the expert roles are designed to approximate these interaction types. 

During training, each expert is evaluated on the original input as well as on modality-replacement inputs where one modality is replaced with random noise. This allows the model to test which modalities each expert depends on. For a uniqueness expert associated with modality $m$, the expert output should remain similar when other modalities are replaced but change when modality $m$ itself is replaced. This encourages the expert to rely primarily on its associated modality. The synergy expert is encouraged to respond to disruptions in cross-modal interactions, while the redundancy expert is encouraged to remain stable when a single modality is replaced, capturing information shared across modalities.

These objectives promote structural role differentiation but do not explicitly enforce separation between expert representations.

\subsection{Representation-level Decorrelation}
To control representation overlap without changing the architecture or routing mechanism, we regularize the anchor latent representations of the experts. The anchor representation refers to the latent representation produced by each expert on the original input sample (i.e., without modality replacement), which serves as the reference representation for decorrelation. Let $\mathbf{z}_e^{(b)} \in \mathbb{R}^{D}$ denote the latent vector produced by expert $e$ for sample $b$ in a batch of size $B$, and let $\mathbf{Z}_e \in \mathbb{R}^{B \times D}$ denote the matrix whose $b$-th row is $\mathbf{z}_e^{(b)}$. Let $P=\{(i,j)\mid i<j\}$ denote expert pairs. We study three complementary objectives that target different notions of inter-expert similarity: 1) sample-wise directional alignment, 2) batch-level representation-space alignment, and 3) cross-feature redundancy. These objectives are chosen because they operate at different levels of representation geometry. For example, Cosine similarity minimization captures instance-level directional alignment between expert representations. CKA-based decorrelation measures population-level similarity between representation spaces, providing a scale-invariant structural comparison. The Barlow-style objective suppresses cross-feature correlation and promotes dimension-level separation. Together, these objectives provide complementary perspectives on inter-expert similarity and enable analysis of how representation separation affects explanation faithfulness.

\subsubsection{Cosine-based Representation Decorrelation}
Let $\hat{\mathbf{z}}_e^{(b)} = \mathbf{z}_e^{(b)} / \|\mathbf{z}_e^{(b)}\|$ denote the $\ell_2$-normalized latent representation of expert $e$ for the $b$-th sample in a batch of size $B$. We minimize squared cosine similarity between expert representations~\cite{rodriguez2016regularizing}, which we refer to as \emph{Rep-Cos}:

\begin{equation}
\mathcal{L}_{\text{cos}}
=
\frac{1}{|P|} \sum_{i<j} \frac{1}{B} \sum_{b=1}^{B}
\left( \langle \hat{\mathbf{z}}_i^{(b)}, \hat{\mathbf{z}}_j^{(b)} \rangle \right)^2.
\end{equation}

\subsubsection{CKA-based Representation Decorrelation}
We measure similarity between expert latent spaces using linear CKA~\cite{kornblith2019similarity,do2025simsmoe}.
Before computing similarity, we center the representations along the batch dimension: \[\widetilde{\mathbf{Z}}_e = \mathbf{H} \mathbf{Z}_e, \qquad \mathbf{H} = \mathbf{I}_B-\frac{1}{B}\mathbf{1}\mathbf{1}^{\top},\] where $\mathbf{H}$ is the centering matrix.

After centering along the batch dimension, linear CKA is computed as \begin{equation}
\text{CKA}(\mathbf{Z}_i, \mathbf{Z}_j)
=
\frac{\|\widetilde{\mathbf{Z}}_i^{\top}\widetilde{\mathbf{Z}}_j\|_F^2}
{\|\widetilde{\mathbf{Z}}_i^{\top}\widetilde{\mathbf{Z}}_i\|_F \|\widetilde{\mathbf{Z}}_j^{\top}\widetilde{\mathbf{Z}}_j\|_F}.
\end{equation} The CKA-based representation decorrelation loss is then defined as \begin{equation}
\mathcal{L}_{\text{cka}}
=
\frac{1}{|P|}
\sum_{i<j}
\text{CKA}(\mathbf{Z}_i, \mathbf{Z}_j).
\end{equation}
We denote this variant as \emph{Rep-CKA}.

\subsubsection{Barlow-style Representation Decorrelation}
We also consider a Barlow-style redundancy reduction objective inspired by Barlow Twins~\cite{zbontar2021barlow}. First, each latent dimension is standardized over the batch: 
\begin{equation}
\bar{z}_{e,a}^{(b)}=\frac{z_{e,a}^{(b)}-\mu_{e,a}}{\sigma_{e,a}+\epsilon},\end{equation} where $\mu_{e,a}$ and $\sigma_{e,a}$ denote the batch mean and standard deviation of feature dimension $a$ for expert $e$, respectively, and $\epsilon>0$ is a small constant for numerical stability. Let $\bar{\mathbf{Z}}_e \in \mathbb{R}^{B\times D}$ denote the standardized latent matrix. For each expert pair $(i,j)$, we compute the cross-correlation matrix
\begin{equation}
\mathbf{C}_{ij} = \frac{1}{B}\bar{\mathbf{Z}}_i^{\top}\bar{\mathbf{Z}}_j \in \mathbb{R}^{D \times D}.
\end{equation} We minimize the off-diagonal entries of this matrix:
\begin{equation}
\mathcal{L}_{\text{barlow}} = \frac{1}{|P|} \sum_{i<j} \frac{1}{D(D-1)} \sum_{a \neq b} \mathbf{C}_{ij}(a,b)^2.\end{equation} This variant is denoted \emph{Rep-Barlow}. Unlike Rep-Cos and Rep-CKA, this objective specifically suppresses cross-feature correlation between experts rather than full representation-space similarity.

All three objectives leave routing and attribution procedures unchanged.

\subsection{Training Objective}
The overall training objective combines task loss, role-based interaction loss, and representation-level decorrelation: \begin{equation}
\mathcal{L} =
\mathcal{L}_{\text{task}} +
\alpha\,\mathcal{L}_{\text{interaction}} +
\lambda\,\mathcal{L}_{\text{decor}}.
\label{eq:train_obj}
\end{equation}
$\mathcal{L}_{\text{task}}$ denotes dataset-specific classification loss, $\mathcal{L}_{\text{interaction}}$ aggregates the role-specific interaction losses, and $\mathcal{L}_{\text{decor}}$ corresponds to one of Rep-Cos, Rep-CKA, or Rep-Barlow. The hyperparameters $\alpha$ and $\lambda$ control the strengths of the interaction and decorrelation terms, respectively.


\section{Experiments}
We study whether reducing inter-expert representation overlap improves explanation faithfulness while preserving predictive performance. Our experiments address four questions. Q1 asks whether representation separation harms task performance. Q2 asks whether reduced overlap improves perturbation-based faithfulness. Q3 asks whether the proposed regularization yields clearer expert specialization in representation geometry and routing behavior. Q4 asks whether the same principle extends beyond role-based MoE architectures.

\subsection{Experimental Setup}
\subsubsection{Datasets}
We evaluate on three multimodal benchmarks. \textbf{ENRICO}~\cite{deka2017rico,liang2021multibench} is a bimodal user interface dataset with screenshot and wireframe modalities, formulated as multiclass classification. \textbf{MIMIC-IV}~\cite{johnson2023mimic} is a trimodal clinical dataset with laboratory measurements, clinical notes, and diagnostic codes; we formulate the task as binary mortality prediction. \textbf{MMIMDb}~\cite{arevalo2017gated} is a bimodal movie genre dataset with textual plot summaries and visual posters, formulated as multi-label classification.

\subsubsection{Training and Optimization}
All models are trained with Adam. To ensure controlled comparison, we freeze pretrained backbone encoders and optimize only lightweight projection layers, the FL-I$^2$MoE experts, the routing MLP, and the prediction heads. The training objective follows Eq.~\eqref{eq:train_obj} and combines task loss, role interaction loss, and representation decorrelation. The task loss is dataset-specific. We use class-weighted cross-entropy for MIMIC-IV, standard cross-entropy for ENRICO, and binary cross-entropy with logits for MMIMDb. 
For each dataset, we adopt the recommended I$^2$MoE hyperparameter configuration~\cite{pmlr-v267-xin25c}, including the learning rate, routing temperature, interaction loss weight, and batch size. Detailed hyperparameter values for each dataset are provided in the supplementary material. We vary only the decorrelation coefficient $\lambda$ in order to isolate the effect of representation separation.

We sweep $\lambda \in \{10^{-5}, 3\times10^{-5}, 10^{-4}, 3\times10^{-4}, 10^{-3}, 3\times10^{-3}, 10^{-2}, 5\times10^{-2}, 10^{-1}, 0.2, 0.5, 1, 2, 3, 4, 6, 8, 10\}$. This grid covers both weak and strong regularization regimes. Values larger than 10 consistently degraded validation performance and were not considered. All experiments are repeated with three random seeds. We report mean and standard deviation for the main predictive and faithfulness results (Tables~\ref{tab:task_performance} and ~\ref{tab:faithfulness_results}). For the remaining analyses, we report mean values only for space and readability. 

\subsubsection{Regularization Variants}
We evaluate three decorrelation strategies: cosine similarity decorrelation (Rep-Cos), CKA-based decorrelation (Rep-CKA), and Barlow-style decorrelation (Rep-Barlow). All variants share the same backbone, optimization settings, and interaction losses. The baseline corresponds to $\lambda=0$.

\subsubsection{Faithfulness Evaluation Protocol}
We evaluate explanation faithfulness using perturbation-based metrics~\cite{alvarez2018robustness,samek2016evaluating,lyu2024towards}. The attribution target corresponds to the model prediction on the original input, so that explanations are evaluated against the evidence actually used by the model.

For a given input $x$, FL-I$^2$MoE produces expert-specific logits $\{\mathbf{y}_e(x)\}_{e=1}^{E}$ and routing weights $\mathbf{w}(x)$. The final logit vector is: \[
\hat{\mathbf{y}}(x) = \sum_{e=1}^{E} w_e(x)\,\mathbf{y}_e(x),
\] and attribution is computed with respect to the predicted class $\hat{y}$.

On the validation split, we compared attention rollout (AttnRoll)~\cite{abnar-zuidema-2020-quantifying}, attention rollout$\times$gradient (AttnRoll$\times$Grad)~\cite{chefer2021generic}, and integrated gradients (IG)~\cite{sundararajan2017axiomatic}. All attribution methods show the same overall direction of change; we report AttnRoll$\times$Grad in the main paper since it was the most stable across datasets and regularization variants on validation. Detailed attribution comparisons are provided in supplementary material.

For gradient-based attribution, we compute gradients of the target logit with respect to expert-specific attention tensors. Since routing weights are derived from shared encoder features and do not depend on expert-internal attention, these gradients already capture the routing effect. We therefore aggregate expert relevance maps across experts for each modality and normalize after aggregation. In contrast, plain AttnRoll is target-agnostic and combines expert maps using routing weights as a heuristic. Features are ranked independently within each modality, and the top-$K\%$ are perturbed using feature-level occlusion~\cite{lyu2024towards}.


Following ERASER-style evaluation~\cite{deyoung2020eraser}, we measure \emph{comprehensiveness} and \emph{sufficiency}: \[
\mathrm{Comp}(x)=\hat y_t(x)-\hat y_t(x_{\mathrm{masked}}),\qquad
\mathrm{Suff}(x)=\hat y_t(x)-\hat y_t(x_{\mathrm{kept}}).
\]
where $x_{\text{masked}}$ denotes the input with the top-$K\%$ most important features replaced by the feature mean, and $x_{\text{kept}}$ denotes the input retaining only the top-$K\%$ most important features, with all remaining features replaced by the feature mean.
We summarize perturbation behavior across $K\in\{5,10,15,20,25\}$ using Area Over the Perturbation Curve (AOPC).

\subsubsection{Validation-based Model Selection}
For each dataset and regularization variant, we select $\lambda$ on the validation split using a two-step procedure. We first retain only those configurations whose Micro-F1/Accuracy and Macro-F1 both remain within an absolute drop of 2\% from the corresponding baseline. Second, among these performance-preserving candidates, we select the $\lambda$ that yields the largest validation AOPC comprehensiveness gap between identified masking and random masking. All performance, faithfulness, and geometry results are evaluated on held-out test data using the checkpoints trained with the selected $\lambda$.

In addition to this selection rule, we track linear CKA as a shared overlap metric because it provides a scale-invariant, population-level measure of inter-expert similarity that is independent of any individual regularizer's objective. This metric is used to analyze how inter-expert representation overlap changes with $\lambda$ and how those changes relate to faithfulness. Note that Rep-CKA directly minimizes CKA, which may give it an advantage under this overlap metric. However, the final  model selection criterion is based on the validation faithfulness gap rather than CKA itself, which reduces the impact of this potential bias.

\subsection{Experimental Results}
We evaluate whether representation-level decorrelation improves explanation faithfulness while preserving predictive performance. 

\subsubsection{Validation Sweeps and Selected Regularization Regimes}
\begin{figure}[tb]
\centering
\resizebox{\textwidth}{!}{%
\begin{tikzpicture}

\begin{groupplot}[
group style={
    group size=2 by 1,
    horizontal sep=1.5cm
},
width=0.5\linewidth,
height=4.5cm,
xmode=log,
xmin=1e-5, xmax=10,
xtick={1e-5,1e-3,1e-1,1,10},
xticklabels={$10^{-5}$,$10^{-3}$,$10^{-1}$,$10^0$,$10^1$},
grid=both,
axis lines=box,
tick label style={font=\scriptsize},
label style={font=\scriptsize},
title style={font=\scriptsize},
legend style={
    font=\scriptsize,
    draw=black,
    fill=white,
    legend columns=3
},
]

\nextgroupplot[
title={Pairwise CKA vs $\lambda$},
xlabel={$\lambda$},
ylabel={Pairwise CKA},
ymin=0, ymax=0.7,
legend to name=sharedlegend
]

\addplot[color=teal, mark=*, mark size=1pt, thick] coordinates{
(1e-5,0.533) (1e-4,0.531) (3e-4,0.538) (1e-3,0.510)
(3e-3,0.494) (1e-2,0.536) (5e-2,0.483) (1e-1,0.340)
(0.2,0.282) (0.5,0.316) (1,0.381) (2,0.522)
(3,0.541) (4,0.568) (6,0.592) (8,0.655) (10,0.652)
};
\addlegendentry{Rep-Cos}

\addplot[color=blue, mark=square*, mark size=1pt, thick] coordinates{
(1e-5,0.482) (1e-4,0.497) (3e-4,0.526) (1e-3,0.501)
(3e-3,0.486) (1e-2,0.386) (5e-2,0.142) (1e-1,0.084)
(0.2,0.041) (0.5,0.016) (1,0.0096) (2,0.0083)
(3,0.0064) (4,0.0045) (6,0.0031) (8,0.0041) (10,0.0040)
};
\addlegendentry{Rep-CKA}

\addplot[color=orange, mark=triangle*, mark size=1pt, thick] coordinates{
(1e-5,0.511) (1e-4,0.522) (3e-4,0.549) (1e-3,0.515)
(3e-3,0.486) (1e-2,0.545) (5e-2,0.487) (1e-1,0.344)
(0.2,0.294) (0.5,0.373) (1,0.480) (2,0.458)
(3,0.583) (4,0.559) (6,0.597) (8,0.577) (10,0.615)
};
\addlegendentry{Rep-Barlow}

\nextgroupplot[
title={Faithfulness gap vs $\lambda$},
xlabel={$\lambda$},
ylabel={Faithfulness gap (\%)},
ymin=0.12, ymax=0.40,
ytick={0.20,0.40},
yticklabels={20,40},
grid=major
]

\addplot[color=teal, mark=*, mark size=1pt, thick] coordinates{
(1e-5,0.162) (1e-4,0.163) (3e-4,0.178) (1e-3,0.159)
(3e-3,0.177) (1e-2,0.151) (5e-2,0.144) (1e-1,0.177)
(0.2,0.224) (0.5,0.190) (1,0.174) (2,0.189)
(3,0.153) (4,0.195) (6,0.212) (8,0.354) (10,0.297)
};

\addplot[color=blue, mark=square*, mark size=1pt, thick] coordinates{
(1e-5,0.209) (1e-4,0.174) (3e-4,0.148) (1e-3,0.185)
(3e-3,0.149) (1e-2,0.162) (5e-2,0.154) (1e-1,0.163)
(0.2,0.251) (0.5,0.226) (1,0.256) (2,0.329)
(3,0.294) (4,0.322) (6,0.288) (8,0.269) (10,0.203)
};

\addplot[color=orange, mark=triangle*, mark size=1pt, thick] coordinates{
(1e-5,0.174) (1e-4,0.144) (3e-4,0.164) (1e-3,0.177)
(3e-3,0.165) (1e-2,0.151) (5e-2,0.152) (1e-1,0.235)
(0.2,0.200) (0.5,0.167) (1,0.139) (2,0.161)
(3,0.186) (4,0.171) (6,0.230) (8,0.264) (10,0.196)
};

\end{groupplot}

\node[font=\small] at ($(group c1r1.south)!0.5!(group c2r1.south)+(0,-1.3cm)$) {\ref{sharedlegend}};

\end{tikzpicture}
}

\caption{
Validation sweeps on ENRICO. Left: mean pairwise linear CKA between expert representations as a function of $\lambda$. Right: validation AOPC comprehensiveness gap between identified and random masking. Moderate regularization reduces overlap and improves faithfulness, while excessively large $\lambda$ can lead to unstable representations.
}

\label{fig:lambda_selection}

\end{figure}
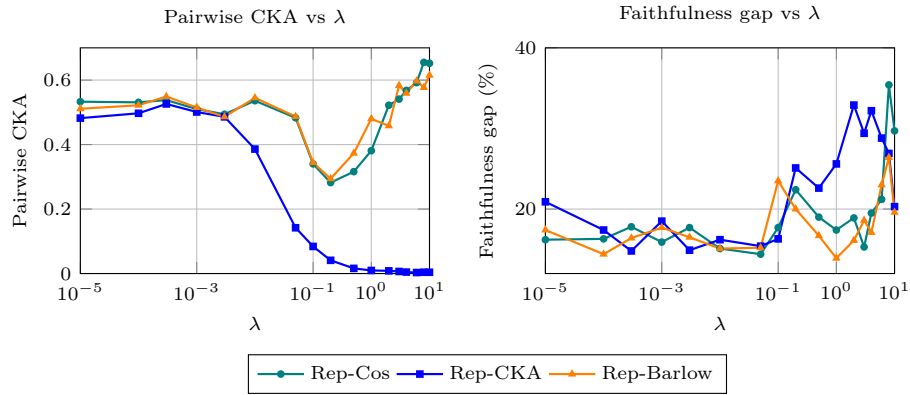

Figure~\ref{fig:lambda_selection} shows validation sweeps on ENRICO as a representative example. The left panel plots the mean pairwise CKA~\cite{kornblith2019similarity} between expert latent representations as a measure of inter-expert overlap, while the right panel shows the validation AOPC comprehensiveness gap between identified and random masking, used as the faithfulness criterion after applying the performance constraint.

Across methods, increasing $\lambda$ does not produce a strictly monotonic change in overlap, but moderate regularization generally reduces inter-expert similarity. The left panel indicates that regularization acts as an effective control knob for overlap. As expected, Rep-CKA shows the most monotonic decrease under the pairwise CKA metric, since its objective directly penalizes CKA between expert latent spaces. We interpret the left panel primarily as a manipulation check rather than a ranking of regularizers. Faithfulness improves mainly in method-dependent intermediate regimes rather than monotonically with increasing $\lambda$ (right panel). In particular, Rep-Cos and Rep-Barlow show a substantial rebound in pairwise CKA at very large $\lambda$, suggesting over-regularization and unstable geometry. These results motivate selecting $\lambda$ from moderate, performance-preserving regimes with reduced overlap rather than simply choosing the strongest regularization.

\subsubsection{Task Performance Preservation (Q1)}
We first examine whether representation-level decorrelation affects predictive performance. Table~\ref{tab:task_performance} reports classification results for the baseline and the three decorrelation variants across datasets. The selected regularized models remain within a pre-specified 2\% absolute margin from the baseline in mean performance across three seeds, indicating that faithfulness improvements are achieved without meaningful degradation in predictive performance. 

\begin{table}[t]
\centering
\footnotesize
\renewcommand{\arraystretch}{0.9}
\caption{Predictive performance across datasets. Results are reported as mean $\pm$ standard deviation over three random seeds. Higher is better for all metrics.}
\label{tab:task_performance}
\resizebox{\linewidth}{!}{
\begin{tabular*}{\linewidth}{@{\extracolsep{\fill}}llcccc}
\toprule
Dataset & Metric & Baseline & Rep-Cos & Rep-CKA & Rep-Barlow \\
\midrule

\multirow{2}{*}{ENRICO}
& Micro-F1  & 52.74$\pm$1.23 & 52.97$\pm$2.06 & 52.05$\pm$1.57 & 51.71$\pm$2.93 \\
& Macro-F1  & 38.50$\pm$1.65 & 38.77$\pm$4.76 & 38.58$\pm$3.53 & 36.83$\pm$5.23 \\

\midrule

\multirow{2}{*}{MIMIC}
& Accuracy & 86.45$\pm$0.30 & 86.18$\pm$0.62 & 86.32$\pm$0.15 & 86.02$\pm$0.74 \\
& Macro-F1 & 81.90$\pm$0.34 & 81.58$\pm$0.98 & 81.84$\pm$0.21 & 80.80$\pm$0.83 \\

\midrule

\multirow{2}{*}{MMIMDb}
& Micro-F1  & 67.90$\pm$0.27 & 66.87$\pm$0.84 & 67.04$\pm$0.30 & 66.59$\pm$0.60 \\
& Macro-F1  & 56.58$\pm$0.51 & 56.08$\pm$0.59 & 56.51$\pm$0.37 & 55.30$\pm$0.35 \\

\bottomrule
\end{tabular*}
}
\end{table}

\subsubsection{Faithfulness Evaluation (Q2)}
\begin{figure*}[t]
\centering
\resizebox{\textwidth}{!}{%
\begin{tikzpicture}
\begin{groupplot}[
group style={
group size=3 by 2,
horizontal sep=1.6cm,
vertical sep=0.6cm
},
width=0.34\textwidth,
height=3.8cm,
xmin=5, xmax=25,
xtick={5,10,15,20,25},
grid=both,
axis lines=box,
tick label style={font=\scriptsize},
label style={font=\scriptsize},
title style={font=\scriptsize},
legend style={
font=\scriptsize,
legend columns=4,
draw=black,
fill=white,
},
]

\nextgroupplot[
title={ENRICO},
ylabel={Comprehensiveness},
legend to name=sharedlegend
]
\addplot[black, thick] coordinates {(5,6.90) (10,13.27) (15,19.30) (20,25.28) (25,30.32)};
\addlegendentry{Baseline}
\addplot[blue, thick] coordinates {(5,12.76) (10,21.05) (15,28.38) (20,34.71) (25,39.54)};
\addlegendentry{Rep-Cos}
\addplot[red, thick] coordinates {(5,18.70) (10,29.10) (15,36.87) (20,43.49) (25,48.10)};
\addlegendentry{Rep-CKA}
\addplot[green!60!black, thick] coordinates {(5,13.44) (10,21.97) (15,29.00) (20,35.19) (25,39.81)};
\addlegendentry{Rep-Barlow}

\nextgroupplot[title={MIMIC}]
\addplot[black, thick] coordinates {(5,2.92) (10,3.82) (15,4.73) (20,5.51) (25,6.26)};
\addplot[blue, thick] coordinates {(5,5.73) (10,6.55) (15,6.53) (20,7.50) (25,8.15)};
\addplot[red, thick] coordinates {(5,6.88) (10,8.77) (15,10.43) (20,11.42) (25,12.36)};
\addplot[green!60!black, thick] coordinates {(5,6.73) (10,6.27) (15,6.20) (20,6.37) (25,6.80)};

\nextgroupplot[title={MMIMDb}]
\addplot[black, thick] coordinates {(5,12.11) (10,18.58) (15,23.72) (20,27.82) (25,31.24)};
\addplot[blue, thick] coordinates {(5,17.03) (10,24.99) (15,30.99) (20,35.45) (25,38.96)};
\addplot[red, thick] coordinates {(5,15.19) (10,22.67) (15,28.39) (20,32.90) (25,36.49)};
\addplot[green!60!black, thick] coordinates {(5,15.20) (10,22.94) (15,28.90) (20,33.52) (25,37.13)};

\nextgroupplot[
ylabel={Sufficiency},
xlabel={Mask ratio (\%)}
]
\addplot[black, thick] coordinates {(5,56.33) (10,51.34) (15,45.93) (20,40.46) (25,35.28)};
\addplot[blue, thick] coordinates {(5,47.77) (10,38.71) (15,30.91) (20,23.78) (25,18.45)};
\addplot[red, thick] coordinates {(5,64.94) (10,55.44) (15,46.48) (20,38.15) (25,31.96)};
\addplot[green!60!black, thick] coordinates {(5,55.71) (10,44.98) (15,35.82) (20,28.91) (25,23.85)};

\nextgroupplot[xlabel={Mask ratio (\%)}]
\addplot[black, thick] coordinates {(5,12.88) (10,11.63) (15,9.83) (20,8.99) (25,7.89)};
\addplot[blue, thick] coordinates {(5,10.14) (10,9.76) (15,9.16) (20,7.67) (25,6.81)};
\addplot[red, thick] coordinates {(5,7.53) (10,6.78) (15,5.51) (20,4.70) (25,3.94)};
\addplot[green!60!black, thick] coordinates {(5,7.34) (10,7.83) (15,7.55) (20,7.23) (25,6.83)};

\nextgroupplot[xlabel={Mask ratio (\%)}]
\addplot[black, thick] coordinates {(5,13.82) (10,9.52) (15,7.19) (20,5.62) (25,4.45)};
\addplot[blue, thick] coordinates {(5,6.79) (10,3.97) (15,2.69) (20,1.88) (25,1.28)};
\addplot[red, thick] coordinates {(5,10.78) (10,6.89) (15,4.94) (20,3.71) (25,2.82)};
\addplot[green!60!black, thick] coordinates {(5,13.34) (10,8.89) (15,6.47) (20,4.87) (25,3.70)};

\end{groupplot}

\node[font=\scriptsize] at ($(group c1r2.south)!0.5!(group c3r2.south)+(0,-1.15cm)$) {\ref{sharedlegend}};

\end{tikzpicture}
}
\caption{Faithfulness perturbation curves using AttnRoll$\times$Grad attribution. Top row shows comprehensiveness and bottom row shows sufficiency across masking ratios $K\in\{5,10,15,20,25\}$. Values denote mean scores across three runs.}
\label{fig:faithfulness_curves}
\end{figure*}
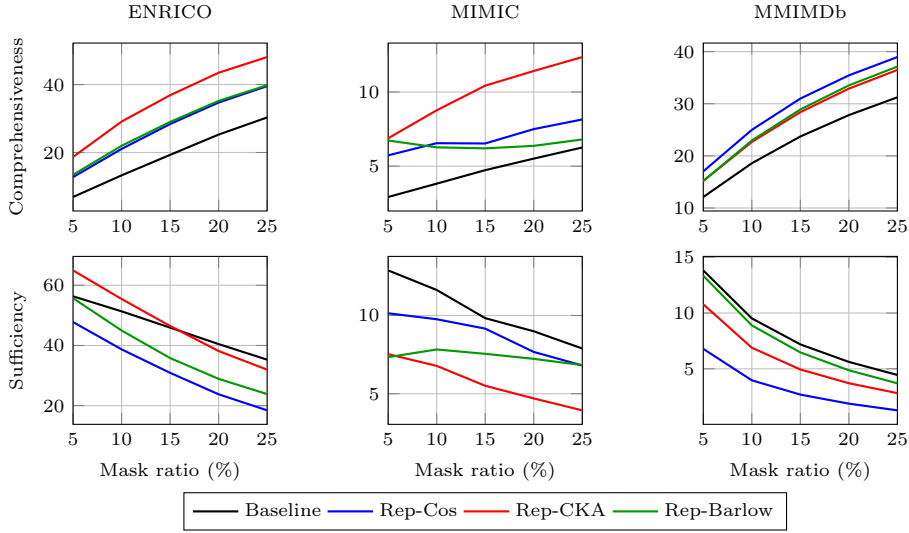

We evaluate whether reducing inter-expert representation overlap improves explanation faithfulness. Figure~\ref{fig:faithfulness_curves} presents perturbation curves for comprehensiveness and sufficiency across masking ratios, while Table~\ref{tab:faithfulness_results} summarizes the corresponding AOPC scores. In Table~\ref{tab:faithfulness_results}, we report identified masking (features selected by the explanation) and random masking as a baseline. Random masking provides a baseline perturbation effect that is independent of the explanation method. The difference between identified and random masking therefore reflects how effectively the attribution method identifies features that truly influence the model prediction. For comprehensiveness, more faithful explanations should yield larger prediction drops under identified masking than under random masking; for sufficiency, they should yield smaller prediction changes when only identified features are retained. Accordingly, we report the identified-minus-random gap for comprehensiveness and the random-minus-identified gap for sufficiency, so that larger gaps indicate that the explanation identifies features with stronger causal influence on the model prediction than randomly selected features.

Across datasets, the decorrelation variants consistently increase the gap between identified and random masking for both metrics, indicating that the explanations better capture decision-relevant evidence that is both necessary and close to minimally sufficient for the model's prediction.
Although some sufficiency curves in Figure~\ref{fig:faithfulness_curves} may visually appear comparable to the baseline, for clarity the curves show only identified masking. When compared against the corresponding random masking values reported in Table~\ref{tab:faithfulness_results}, the decorrelation variants consistently yield stronger perturbation effects than the baseline. 
Overall, these results support our hypothesis that reducing inter-expert representation overlap leads to more faithful feature-level explanations.

\begin{table}[t]
\centering
\footnotesize
\renewcommand{\arraystretch}{0.9}
\caption{Faithfulness evaluation using AOPC metrics.
Each entry is reported as gap on the first line and random / identified on the second line.
For comprehensiveness, gap = identified $-$ random.
For sufficiency, gap = random $-$ identified (higher is better).}
\label{tab:faithfulness_results}
\resizebox{\linewidth}{!}{
\begin{tabular*}{\linewidth}{@{\extracolsep{\fill}}llcccc}
\toprule
Dataset & Metric & Baseline & Rep-Cos & Rep-CKA & Rep-Barlow \\
\midrule

\multirow{2}{*}{ENRICO}
& \raisebox{6pt}[0pt][0pt]{AOPC Comp gap$\uparrow$}
& \shortstack{13.21$\pm$2.94 \\ \scriptsize(5.80 / 19.01)}
& \shortstack{22.39$\pm$4.60 \\ \scriptsize(4.90 / 27.29)}
& \shortstack{\textbf{29.05$\pm$11.44} \\ \scriptsize(6.20 / 35.22)}
& \shortstack{22.68$\pm$8.78 \\ \scriptsize(5.20 / 27.88)} \\[2pt]

& \raisebox{6pt}[0pt][0pt]{AOPC Suff gap$\uparrow$}
& \shortstack{8.76$\pm$2.53 \\ \scriptsize(54.63 / 45.87)}
& \shortstack{\textbf{23.03$\pm$5.71} \\ \scriptsize(54.95 / 31.92)}
& \shortstack{16.17$\pm$10.44 \\ \scriptsize(63.57 / 47.39)}
& \shortstack{19.75$\pm$7.47 \\ \scriptsize(57.60 / 37.86)} \\

\midrule

\multirow{2}{*}{MIMIC}
& \raisebox{6pt}[0pt][0pt]{AOPC Comp gap$\uparrow$}
& \shortstack{3.31$\pm$1.42 \\ \scriptsize(1.33 / 4.65)}
& \shortstack{5.55$\pm$4.04 \\ \scriptsize(1.34 / 6.89)}
& \shortstack{\textbf{8.83$\pm$2.97} \\ \scriptsize(1.14 / 9.97)}
& \shortstack{5.28$\pm$1.27 \\ \scriptsize(1.19 / 6.47)} \\[2pt]

& \raisebox{6pt}[0pt][0pt]{AOPC Suff gap$\uparrow$}
& \shortstack{3.78$\pm$2.15 \\ \scriptsize(14.02 / 10.24)}
& \shortstack{5.47$\pm$3.28 \\ \scriptsize(14.17 / 8.71)}
& \shortstack{\textbf{8.44$\pm$3.25} \\ \scriptsize(14.13 / 5.69)}
& \shortstack{4.72$\pm$2.06 \\ \scriptsize(12.08 / 7.36)} \\

\midrule

\multirow{2}{*}{MMIMDb}
& \raisebox{6pt}[0pt][0pt]{AOPC Comp gap$\uparrow$}
& \shortstack{20.37$\pm$2.59 \\ \scriptsize(2.32 / 22.69)}
& \shortstack{\textbf{27.06$\pm$4.92} \\ \scriptsize(2.43 / 29.48)}
& \shortstack{24.70$\pm$1.20 \\ \scriptsize(2.43 / 27.13)}
& \shortstack{24.79$\pm$2.82 \\ \scriptsize(2.75 / 27.54)} \\[2pt]

& \raisebox{6pt}[0pt][0pt]{AOPC Suff gap$\uparrow$}
& \shortstack{23.89$\pm$2.65 \\ \scriptsize(32.01 / 8.12)}
& \shortstack{26.16$\pm$2.15 \\ \scriptsize(29.48 / 3.32)}
& \shortstack{\textbf{26.50$\pm$0.75} \\ \scriptsize(32.33 / 5.83)}
& \shortstack{26.06$\pm$1.29 \\ \scriptsize(33.51 / 7.45)} \\

\bottomrule
\end{tabular*}
}
\end{table}

\subsubsection{Representation Geometry Analysis (Q3)}
To examine whether representation-level decorrelation leads to clearer expert specialization, we analyze the geometry of expert latent representations. Table~\ref{tab:rep_geometry} reports separation metrics including silhouette score~\cite{9260048}, pairwise CKA, and pairwise correlation, capturing complementary aspects of representation overlap. The silhouette score reflects cluster-level separation, while pairwise CKA and correlation measure subspace similarity and linear redundancy between expert representations.

Across datasets, the decorrelation variants consistently reduce inter-expert similarity as reflected by lower pairwise CKA values. Improvements in silhouette score are also observed in several settings, particularly on ENRICO and MMIMDb, indicating clear cluster-level separation between experts. In contrast, the MIMIC dataset already exhibits relatively high silhouette scores in the baseline, leaving limited room for further improvement. Different regularization strategies influence different aspects of representation geometry. Rep-Cos tends to produce stronger improvements in silhouette score, whereas Rep-CKA and Rep-Barlow more consistently reduce pairwise similarity metrics such as CKA and correlation.

Overall, these results indicate that the proposed regularization encourages experts to occupy more distinct regions of the representation space. This separation aligns with the faithfulness improvements in Table~\ref{tab:faithfulness_results}, supporting our hypothesis that reduced inter-expert overlap contributes to more faithful feature-level explanations. In the following experiment, we further examine whether this separation also improves role alignment under controlled synthetic settings.

\begin{table}[t]
\centering
\footnotesize
\caption{Expert representation geometry under decorrelation regularization.
Each cell reports the selected value with the change from the baseline in parentheses. Higher silhouette
indicates better expert separation, while lower CKA and correlation
indicate reduced representation overlap.}
\label{tab:rep_geometry}
\resizebox{\linewidth}{!}{
\begin{tabular*}{\linewidth}{@{\extracolsep{\fill}}llccc}
\toprule
Dataset & Method (Selected $\lambda)$ & Silhouette $\uparrow$ & Pairwise CKA $\downarrow$ & Pairwise Corr $\downarrow$ \\
\midrule

\multirow{3}{*}{ENRICO}
 & Rep-Cos (0.2)    & 0.605 (+0.544) & 0.206 (-0.274) & 0.421 (-0.169) \\
 & Rep-CKA (0.2)    & 0.231 (+0.171) & 0.050 (-0.430) & 0.159 (-0.430) \\
 & Rep-Barlow (0.1) & 0.248 (+0.188) & 0.234 (-0.246) & 0.453 (-0.136) \\

\midrule

\multirow{3}{*}{MIMIC}
 & Rep-Cos (0.0001)     & 0.791 (-0.023) & 0.326 (-0.108) & 0.332 (+0.039) \\
 & Rep-CKA (0.1)     & 0.729 (-0.085) & 0.303 (-0.13) & 0.252 (-0.041) \\
 & Rep-Barlow (0.0001) & 0.667 (-0.148) & 0.128 (-0.306) & 0.173 (-0.119) \\

\midrule

\multirow{3}{*}{MMIMDb}
 & Rep-Cos (8.0)    & 0.806 (+0.360) & 0.095 (-0.06) & 0.087 (-0.126) \\
 & Rep-CKA (0.0001)    & 0.462 (+0.015) & 0.124 (-0.031) & 0.193 (-0.021) \\
 & Rep-Barlow (4.0) & 0.699 (+0.253) & 0.002 (-0.153) & 0.003 (-0.21) \\

\bottomrule
\end{tabular*}
}
\end{table}

\begin{table}[t]
\centering
\begin{minipage}[t][4.5cm][t]{0.45\textwidth}
\centering
\caption{Role alignment on the synthetic benchmark. Higher values indicate stronger alignment between routing behavior and the target role.}
\label{tab:role_alignment_usage}
\vspace*{\fill}

\footnotesize
\resizebox{\linewidth}{!}{
\begin{tabular}{lccc}
\toprule
Setting & Baseline & Rep-CKA & Selected \\
        & (\%) & (\%) & $\lambda$ \\
\midrule
Uniqueness$_0$ & 30.68 & 41.64 & 10 \\
Uniqueness$_1$ & 22.16 & 33.04 & 10 \\
Synergy        & 16.13 & 27.00 & 3  \\
Redundancy     & 26.15 & 39.61 & 6  \\
\bottomrule
\end{tabular}}
\end{minipage}
\hfill
\begin{minipage}[t][4.5cm][t]{0.50\textwidth}
\centering
\caption{Generalization to sparse MoE (GShardGate) on ENRICO. Faithfulness improves while predictive performance remains comparable.}
\label{tab:gshard_generalization}
\vspace*{\fill}
\footnotesize
\resizebox{0.85\linewidth}{!}{
\begin{tabular}{llcc}
\toprule
& Metric & Baseline & Rep-CKA \\
\midrule
\multirow{2}{*}{Task}
  & Micro-F1 $\uparrow$ & 51.26{\footnotesize$\pm$0.20} & 51.37{\footnotesize$\pm$0.68} \\
  & Macro-F1 $\uparrow$ & 35.73{\footnotesize$\pm$1.20} & 36.33{\footnotesize$\pm$5.26} \\
\midrule
\multirow{4}{*}{Faith.}
  & AOPC Comp & 20.24 & 28.97 \\
  & {\scriptsize(rnd/id) $\uparrow$} & {\scriptsize(5.56/25.80)} & {\scriptsize(4.95/33.92)} \\
  & AOPC Suff & 19.71 & 27.01 \\
  & {\scriptsize(rnd/id) $\uparrow$} & {\scriptsize(54.88/35.17)} & {\scriptsize(54.85/27.83)} \\
\bottomrule
\end{tabular}}
\end{minipage}
\end{table}

\subsubsection{Role Alignment on Synthetic Data (Q3)}
While the previous analysis confirms that representation-level decorrelation reduces inter-expert overlap and aligns with improved faithfulness, we examine whether this separation leads to better alignment between experts and their semantic roles. We use the synthetic benchmark introduced in InterSHAP~\cite{wenderoth2025measuring}, which generates multimodal data with controlled interaction structures. In our two-modality setting, labels are constructed using predefined interaction patterns. Uniqueness settings depend on a single modality, redundancy settings replicate the same signal across modalities, and synergy settings are generated using non-linear interactions such as XOR. Because the underlying interaction type is known, this dataset allows us to directly evaluate whether routing concentrates on the corresponding expert. 

We use Rep-CKA in this experiment since it directly optimizes the representation overlap metric used throughout the paper. The goal is not to rank regularizers, but to test whether reducing inter-expert overlap improves routing alignment under known interaction structure. For each setting, we train the model with Rep-CKA and measure target usage, defined as the average routing weight assigned to the expert associated with the ground-truth interaction type.  

Table~\ref{tab:role_alignment_usage} shows that representation-level separation increases routing concentration on the correct expert. This result provides controlled evidence that reducing representation overlap strengthens semantic role specialization.


\subsubsection{Generalization Beyond Role-based MoE (Q4)}
We test whether representation-level separation improves explanation faithfulness in standard sparse MoE architectures that do not explicitly encode semantic expert roles. Specifically, we apply Rep-CKA regularization to a Transformer model equipped with GShardGate~\cite{lepikhin2020gshard} routing and evaluate it on the ENRICO dataset using the same feature-level perturbation protocol as in the previous experiments. Results on MIMIC and MMIMDb show similar trends and are reported in the supplementary material.

Table~\ref{tab:gshard_generalization} reports the results. While predictive performance remains largely unchanged compared to the baseline, Rep-CKA increases both the comprehensiveness gap and the sufficiency gap. These results suggest that representation-level decorrelation is not limited to role-based architectures such as I$^2$MoE. Even in a standard sparse MoE baseline without explicit role decomposition, reducing overlap improves explanation faithfulness. 

\section{Ablation Study}
We perform a component ablation to examine the roles of interaction learning and representation decorrelation. In particular, we compute four settings: (1) no interaction loss and no CKA decorrelation, (2) interaction loss only, (3) CKA decorrelation only, and (4) the full model with both components.

Table~\ref{tab:ablation_interaction_cka} reports predictive performance and faithfulness metrics on ENRICO. The interaction-only model achieves similar predictive performance but does not improve faithfulness, indicating that structural role assignments alone are insufficient when expert representations remain highly overlapping. The Rep-CKA-only model provides a partial benefit, improving comprehensiveness but remaining well below the full model and not uniformly improving all faithfulness metrics relative to Base MoE. The largest gains are obtained when both components are combined, indicating that role specialization becomes most effective when supported by explicit representation separation.

\begin{table}[t]
\centering
\footnotesize
\caption{Ablation study on ENRICO evaluating interaction learning and Rep-CKA decorrelation. Faithfulness is reported using the same gap definitions as in Table~\ref{tab:faithfulness_results}.}
\label{tab:ablation_interaction_cka}
\resizebox{\linewidth}{!}{
\begin{tabular}{c|cc|cc|cc}
\toprule
Method 
& Interaction & Rep-CKA 
& Micro-F1 $\uparrow$ & Macro-F1 $\uparrow$
& AOPC Comp gap $\uparrow$ & AOPC Suff gap $\uparrow$ \\
\midrule
Base MoE & No  & No  & 52.05$\pm$2.74 & 35.07$\pm$2.43 & 16.45 & 13.50\\
I$^2$MoE (Baseline) & Yes & No  & \textbf{52.74$\pm$1.23} & 38.50$\pm$1.65 & 13.21 & 8.76\\
Rep-CKA only & No  & Yes & 52.05$\pm$0.68 & 33.71$\pm$3.12 & 17.53 & 10.01 \\
\rowcolor{gray!15}
Ours & Yes & Yes & 52.05$\pm$1.57 & \textbf{38.58$\pm$3.53} & \textbf{29.05} & \textbf{16.17} \\
\bottomrule
\end{tabular}
}
\end{table}

\section{Conclusion}
We studied whether role specialization in MoE models improves explanation faithfulness and found that structural role assignments alone do not guarantee faithful explanations when expert representations remain highly overlapping. Across multimodal benchmarks, representation-level decorrelation consistently improves perturbation-based faithfulness while preserving predictive performance, and ablation results show that role-based interaction learning is most effective when paired with explicit representation separation. However, the most effective decorrelation objective varies across datasets and faithfulness metrics. This suggests that expert separation alone does not fully determine explanation faithfulness, and its impact depends on additional factors such as modality structure, interaction complexity, and the dynamics of expert specialization. Similar trends observed in standard sparse MoE models further indicate that representation separation may be useful beyond explicitly role-based architectures.

\begin{credits}
\subsubsection{\ackname} This research was  supported by the Alberta Machine Intelligence Institute (Amii), NSERC (including grants DGECR-2022-00369, RGPIN-2022-03469, and RGPIN-2025-05572), and Alberta Innovates (Enabling Better Health through Artificial Intelligence (AI-Better Health) Program).

\subsubsection{\discintname}
The authors have no competing interests to declare 
that are relevant to the content of this article.
\end{credits}
%
%
%
\bibliographystyle{splncs04}
%

\bibliography{reference}

\end{document}